# A Symbolic Approach to Reasoning with Linguistic Quantifiers

**Didier Dubois**
**Henri Prade**
Institut de Recherche en Informatique de Toulouse
C.N.R.S. & Université Paul Sabatier
118 route de Narbonne
31062 Toulouse Cedex, France

**Lluis Godo**
**Ramon López de Màntaras**
Institut d'Investigacio en Intelligencia Artificial
CEAB-CSIC
Camí de Santa Bàrbara
17300 Blanes, Spain

## Abstract

This paper investigates the possibility of performing automated reasoning in probabilistic logic when probabilities are expressed by means of linguistic quantifiers. Each linguistic term is expressed as a prescribed interval of proportions. Then instead of propagating numbers, qualitative terms are propagated in accordance with the numerical interpretation of these terms. The quantified syllogism, modelling the chaining of probabilistic rules, is studied in this context. It is shown that a qualitative counterpart of this syllogism makes sense, and is relatively independent of the threshold defining the linguistically meaningful intervals, provided that these threshold values remain in accordance with the intuition. The inference power is less than that of a full-fledged probabilistic constraint propagation device but better corresponds to what could be thought of as commonsense probabilistic reasoning.

## 1 INTRODUCTION

Precise values of probabilities are not always available. Experts often assess probabilities under the form of intervals (e.g. "between 80 and 90 % of A's are B's") or even linguistically (e.g. "almost all A's are B's"), or are only able to rank-order probability values, stating that a probability is certainly greater than another. Thus it raises the question of the possibility of reasoning with probabilities in a qualitative way. The main appeal of a qualitative approach (when such an approach is feasible), is that it requires less precision than a pure numerical representation while still leading to meaningful conclusions in the reasoning process. Also, the qualitative approach allows us to have a better interface with human users, in a way more compatible with their own reasoning processes. The idea of reasoning qualitatively with probabilities has been investigated along different lines by various researchers in Artificial Intelligence especially in the last five years. A first family of approaches works with inequalities between probabilities (e.g.Wellman (1990)). A second type of approach considers probability-like functions which take their values in a finite totally ordered set not related to [0,1] (e.g.Yang, Beddoes and Poole, 1990). Another kind of qualitative probability approach is Adams (1975)' conditional logic (see also Pearl (1988)) which manipulates infinitesimal probabilities. For the sake of brevity we do not mention other logical approaches to probabilities here.

The approach developed in this paper maintains an interpretation of qualitative (linguistic) probability values in terms of numerical intervals. Here, linguistic quantifiers such as most, few, etc... are viewed as imprecisely or fuzzily known conditional probabilities, i.e. terms represented by crisp, or in the most general case, fuzzy subintervals of [0,1] (Zadeh, 1985 ; Dubois and Prade, 1988). Here, an ordered set of elementary labels of quantifiers is chosen in order to provide a linguistic scale for conditional probabilities (or proportions) used in default rules like "Q A's are B's", where Q is viewed as the answer to the question : "how many A's are B's ?". A qualitative algebra (Q-algebra) (Travé-Massuyès and Piera, 1989) is defined on the set of possible labels, built from the elementary labels forming the scale. Inference rules which are the qualitative counterparts of numerical formulas for computing bounds on probabilities in quantified syllogisms or similar propagation rules, can be proposed for reasoning in qualitative probability networks.

The next section discusses how to build a set of linguistic labels to be used in the qualitative probability computations. Section 3 gives the necessary background about local patterns of inference used to propagate constraints on probabilities known to belong to intervals. Section 4 defines qualitative versions of these rules of inference. Section 5 discusses the robustness of the approach, i.e. to what extent the qualitative calculus remains unchanged when the numerical interpretation of the linguistic labels is slightly modified. A qualitative analysis of inference rules in Adams' probabilistic logic is given in Section 6. Section 7 discusses the problems encountered when trying to develop a qualitative constraint propagation rule based on Bayes theorem. Section 8 gives an example and shows how the constraint propagation-based strategy, recalled in Section 3, to answer queries about conditional probabilities can be adapted to the qualitative setting.

## 2   LATTICES OF LABELS

Let us consider an ordered set of elementary labels of linguistic quantifiers that may account for any probability value. Each label corresponds to a subinterval of the unit interval, and the set of labelled subintervals completely covers it. So a linguistic scale will be made of the labels of a collection of subintervals covering [0,1] of the form {0, (0, $a_1$], [$a_1$, $a_2$], ..., [$a_{n-1}$, $a_n$], [$a_n$, 1), 1}. For convenience we shall call a "partition" such a collection, although the intervals overlap at their edges, except in 0 and 1 which are dealt with separately due to their particular meanings corresponding to 'none' and 'all'.

Let $\mathcal{P}$ be a partition of [0,1] in subintervals representing quantifiers from a linguistic scale. By convention, both the linguistic scale and the corresponding partition will be denoted $\mathcal{P}$. It seems reasonable that this linguistic scale should be symmetric with respect to 0.5 since the antonym of each linguistic quantifier in the scale should also be in the scale. Linguistic antonymy, for instance *ANT(Almost none) = Almost all* or *ANT(Few) = Most*, is expressed at the numerical level by relations like ANT([a,b]) = [1 – b, 1 – a], since intervals are used to represent the meaning of linguistic quantifiers. As a consequence if P(A) is the probability of event A, linguistically qualified by $X \in \mathcal{P}$, then P($\bar{A}$) the probability of the complementary event $\bar{A}$ should be ANT(X) $\in \mathcal{P}$.

The universe of description U induced by a partition $\mathcal{P}$ is defined as the set of intervals that are union of adjacent elements of $\mathcal{P}$. The set inclusion relationship ($\subseteq$) provides U with an ordered structure that has a tree representation. For instance, if we take parameters "a" and "b" to be smaller than 0.5, then [0,1] can be (non strictly) symmetrically partitioned as

$$\mathcal{P} = \{ 0, (0,a], [a, b], [b, 1 - b], [1 - b, 1 - a], [1 - a, 1), 1 \}$$

corresponding to the following linguistic quantifiers :

0 ::= *None*
(0, a] ::= *Almost none* (*Al-none* for short)
[a, b] ::= *Few*
[b, 1 – b] ::= *About half* (*Ab-half* for short)
[1 – b, 1 – a] ::= *Most*
[1 – a, 1) ::= *Almost all* (*Al-all* for short)
1 ::= *All*

The set $\mathcal{P}$ constitutes the highest meaningful level of specificity with respect to the language. Between this level and the least specific one (i.e. the whole interval [0,1]), the universe of description U contains several internal ordered levels of specificity. For example, with the seven terms defined above we have five levels in between ; see Figure 1.

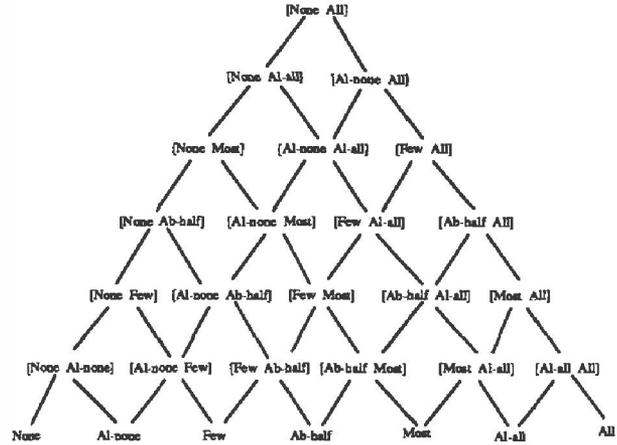

Figure 1 : Specificity ordering

The set of elementary (most specific) linguistic quantifiers can also be ordered according to the usual certainty ordering in the unit interval :

*None ≤ Almost none ≤ Few ≤ About half ≤ Most ≤ Almost all ≤ All*

This ordering enables us to consider higher level elements of the universe U as intervals defined on the partition set, for instance, [Few Most] = {X $\in \mathcal{P}$ | Few ≤ X ≤ Most} and it is fully compatible with the above numerical interpretation in terms of probability intervals. The semantics of the higher level elements of the universe corresponds to the convex hull of the intervals attached to their edges. For instance, the (numerical) interpretation of *[Few Most]*, i.e. 'From few to most', is the interval *[a, 1-a]*. The certainty ordering can be partially extended to the whole universe U as well, by defining for every [$X_1$, $Y_1$], [$X_2$, $Y_2$] $\in$ U

[$X_1$, $Y_1$] ≤ [$X_2$, $Y_2$] if, and only if, $X_1 \leq X_2$ and $Y_1 \leq Y_2$,

giving rise to a structure which differs from the previous one. Such a double ordering structure is in accordance with bilattices as discussed in (Ginsberg, 1988).

## 3   LOCAL PROPAGATION OF INTERVAL-VALUED PROBABILITIES

In (Amarger, Dubois and Prade, 1991b), a local computation approach which deals with interval-valued conditional probabilities is presented. In the approach a basic pattern for local inference is the following so-called 'quantified syllogism' :

P(B|A)$\in$ [P*(B|A),P*(B|A)]; P(A|B)$\in$ [P*(A|B),P*(A|B)]
P(C|B)$\in$ [P*(C|B),P*(C|B)]; P(B|C) $\in$ [P*(B|C),P*(B|C)]

---

P(C|A) = ?          P(A|C) = ?

where $P_*$ and $P^*$ respectively denote lower and upper bounds, and where we want to compute (the tightest) bounds which can be deduced for $P(C|A)$ and $P(A|C)$.

The following bounds have been established in (Dubois and Prade, 1988 ; Dubois, Prade and Toucas, 1990) and have been shown to be the tightest ones when $P(B|A)$, $P(A|B)$, $P(B|C)$ and $P(C|B)$ are *precisely known* (i.e. $P(B|A) = P_*(B|A) = P^*(B|A)$, etc.), and are different from 0 or 1 :

lower bound :

$$P_*(C|A) = P_*(B|A) \max\left(0, 1 - \frac{1 - P_*(C|B)}{P_*(A|B)}\right)$$

upper bound :

$$P^*(C|A) = \min\left(\begin{array}{l} 1, \; 1 - P_*(B|A) + \dfrac{P_*(B|A) \cdot P^*(C|B)}{P_*(A|B)}, \\ \dfrac{P^*(B|A) \cdot P^*(C|B)}{P_*(A|B) \cdot P_*(B|C)}, \\ \dfrac{P^*(B|A) \cdot P^*(C|B)}{P_*(A|B) \cdot P_*(B|C)}[1 - P_*(B|C)] + P^*(B|A) \end{array}\right)$$

Related local patterns of inference for interval-valued conditional probabilities have been independently developed by Güntzer, Kießling and Thöne (1991), Thöne et al. (1991a) and by Heinsohn (1991) in the contexts of deductive data bases and of terminological languages respectively. While the above lower bound is still optimal when only bounds are known on $P(B|A)$, $P(A|B)$ and $P(C|B)$, Thöne, Güntzer and Kießling (1991b) have recently pointed out that the above upper bound can be improved when only lower and upper bounds on the probabilities are available in the syllogism. This is basically due to the fact that the third and fourth terms are linearly increasing with respect to $P(B|A)$ while the second term is linearly decreasing in $P(B|A)$ if $P^*(C|B) < P_*(A|B)$. These authors show that the above upper bound becomes optimal provided that we add the following fifth term in the above minimum of four terms :

$$\frac{P^*(C|B)}{P^*(C|B) + P_*(B|C) \cdot (P_*(A|B) - P^*(C|B))}.$$

This fifth term is simply obtained by computing the value of $P(B|A)$ that makes the second and third term equal. This fifth term does improve the upper bound if and only if $P_*(A|B) > P^*(C|B)$, and moreover the interval $[P_*(B|A), P^*(B|A)]$ contains the quantity

$$\frac{P_*(B|C) \cdot P_*(A|B)}{P_*(B|C) \cdot P_*(A|B) + P^*(C|B) \cdot (1 - P_*(B|C))}.$$

The local inference approach proposed in Amarger et al. (1991b) also takes advantage of an extended form of Bayes rule expressed in terms of conditional probabilities only, namely

$$\forall A_1,\ldots,A_k, \; P(A_1|A_k) = P(A_k|A_1) \prod_{i=1}^{k-1} \frac{P(A_i|A_{i+1})}{P(A_{i+1}|A_i)}$$

(with all involved quantities positive), from which useful inequalities are obtained in the case where only lower and upper bounds are available.

The constraint propagation method which is used is the following : recursively apply the quantified syllogism to generate upper and lower bounds of missing probabilities. This step is performed until the probability intervals can no longer be improved. Then recursively apply the extended Bayes rule to improve the bounds thus generated, and continue the whole procedure until no improvement takes place. This constraint propagation method can sometimes give bounds as tight as the best ones computed by a global optimization method based on linear programming (see Amarger et al., 1991b).

## 4   THE QUALITATIVE QUANTIFIED SYLLOGISM

### 4.1   COMPUTATION OF THE QUALITATIVE TABLE

In this section we will focus on the qualitative counterpart of the quantified syllogism inference pattern, recalled in the preceding section. We use the following notations, where $Q_i$ are linguistic labels.

> Q1 A's are B's ; Q2 B's are A's
> Q3 C's are B's ; Q4 B's are C's
> ___
> Q5 A's are C's ; Q6 C's are A's

This inference rule is interesting from the point of view of commonsense reasoning since it offers a precise model of chaining uncertain "if... then..." statements expressed by means of imprecise quantifiers or conditional probabilities. In the following we build the qualitative functions (Q-functions) corresponding to that pattern, i.e. we build a table giving qualitative values for Q5 and Q6 for any combination of possible qualitative values for Q1, Q2, Q3 and Q4.

The process of building the Q-functions is performed according to the following steps :

1. Consider a linguistic scale of linguistic quantifiers together with a suitable partition of the unit interval [0,1] that represents them. In what follows we will use the partition $\mathcal{P}$ defined above with parameters a = 0.2, and b = 0.4, that is, (0,0.2] = Almost none ; [0.2, 0.4] = Few ; [0.4,0.6] = About half ; [0.6,0.8] = Most ; [0.8,1) = Almost all.

2. Consider all possible combinations of these linguistic values for $P(B|A)$, $P(A|B)$, $P(B|C)$ and $P(C|B)$.

3. For each of such combinations, compute the lower and upper bounds of $P(C|A)$ (and $P(A|C)$) using the numerical expression of the pattern given in Section

3. For example if Q1 = Most, Q2 = Almost all, Q3 = About half, Q4 = Almost all, then :

$P^*(B|A) = 0.8 \quad P_*(B|A) = 0.6$
$P^*(A|B) = 1 \quad P_*(A|B) = 0.8$
$P^*(B|C) = 0.6 \quad P_*(B|C) = 0.4$
$P^*(C|B) = 1 \quad P_*(C|B) = 0.8$

which gives

$P^*(C|A) = 1 \quad P_*(C|A) = 0.45$
$P^*(A|C) = 1 \quad P_*(A|C) = 0.3$.

4. These results are then approximated by means of the most specific element of the universe of description U (see Figure 1) which contains them. So, the interval [0.45, 1] for P(C|A) is approximated to the larger interval [0.4, 1], that is, the resulting Q5 is set to [About-half, All]. In the same way, Q6 is approximated to [Few, All].In this way, we have partially defined the qualitative functions Q5 and Q6, i.e. defined as functions

$$Q5, Q6 : \mathcal{P} \times \mathcal{P} \times \mathcal{P} \times \mathcal{P} \longrightarrow U.$$

5. Finally, the complete definition of the qualitative functions Q5, Q6 : U × U × U × U ------> U can be easily derived from the above partially defined ones by simply applying them on the upper and lower bounds (which are elements of $\mathcal{P}$) of the non-elementary elements of U, and then taking the convex hull. For instance :

Q5([Most, All], All, [None, All], Almost-all) =
= convex_hull_of(Q5(Most, All, None, Almost-all), Q5(All, All, None, Almost-all), Q5(Most, All, All, Almost-all), Q5(All, All, All, Almost-all) )
= convex_hull_of([About-half, All], Almost-all, [About-half, Most], Almost-all) = [About-half, All]

**Remark** : Note that in the above procedure, the qualitative calculation table for the quantified syllogism is computed by using the approximation step only at the end of the computation. Another approach one may think of would be to have precomputed tables for product and quotient, and to use them in the calculation of the bounds. However this latter approach would not be satisfactory because it yields too imprecise results.

### 4.2 THE 5-QUANTIFIER CASE

In this section we analyse the results obtained on the most elementary type of qualitative scale of linguistic quantifiers, i.e. {none, few, about half, most, all} where *few* is of the form [ε,α] for some positive, infinitesimal value ε, α is some number in (0, 1/2), *about half* is interpreted as [α, 1 − α], and most is [1 − α, 1 − ε]. Note that the name "about half" is indeed short for "neither few nor most, but in between", since the interval [α, 1 − α] may be quite imprecise.

Table 1 gives the complete results when α = 0.3 ; the table is sorted by putting together the 4-tuples (Q1 Q2 Q3 Q4) that lead to the same value of Q5. A first remark is that in many situations when none of the quantifiers mean "all", no information is obtained on P(C|A). This is especially true when both P(A|B) and P(B|A) take small qualitative values. Some lines of the table may look surprising. For instance we see that nothing can be inferred from the four statements

"all A's are B's"; "most B's are A's"
"all C's are B's"; "about half of the B's are C's".

Especially, the lower bound $P_*(C|A) = 0$ is attained in this case if pessimistic interpretations of "most" and "about half" are chosen, say 70 % and 30 % respectively.

| P(B\|A) | P(A\|B) | P(B\|C) | P(C\|B) | P(C\|A) |
|---|---|---|---|---|
| none | none | [none,most] | [none,all] | [none,all] |
| few | few | [few,most] | [few,most] | |
| few | half | [few,half] | [few,most] | |
| few | most | [few,half] | half | |
| [half,all] | most | [few,most] | half | |
| [half,all] | [few,half] | [few,most] | [few,most] | |
| all | [few,half] | all | [few,most] | |
| all | most | all | half | |
| few | [few,all] | none | none | [none,most] |
| few | half | most | most | |
| [few,half] | most | few | few | |
| most | [few,half] | all | [few,most] | |
| most | most | all | half | |
| few | half | most | [few,half] | [none,half] |
| few | most | half | few | |
| few | most | most | half | |
| half | [few,half] | all | [few,most] | |
| half | [few,all] | none | none | |
| [most,all] | most | [few,all] | few | |
| few | [few,half] | all | [few,most] | [none,few] |
| few | most | most | few | |
| few | most | all | [few,half] | |
| half | most | all | few | |
| most | [few,all] | none | none | |
| none | none | all | [few,all] | {none,none} |
| all | [few,all] | none | none | |
| few | few | most | all | [few,all] |
| few | [few,all] | [few,half] | all | |
| [few,half] | most | [few,half] | most | |
| half | most | most | most | |
| few | half | most | all | [few,most] |
| few | all | few | [few,most] | |
| few | all | half | most | |
| half | all | [few,half] | [half,most] | |
| half | all | few | few | |
| half | all | most | most | |
| most | all | [few,most] | half | |
| few | [most,all] | most | [most,all] | [few,half] |
| few | all | half | half | |
| half | most | all | most | |
| half | all | half | few | |
| half | all | all | [half,most] | |
| most | all | [few,most] | few | |
| most | all | all | half | |
| few | [few,all] | all | all | [few,few] |
| few | most | all | most | |
| few | all | half | few | |
| few | all | most | [few,half] | |
| few | all | all | [half,most] | |
| half | all | [most,all] | few | |
| most | all | all | few | |
| all | all | [few,all] | few | |
| half | [few,all] | [few,most] | all | [half,all] |
| [most,all] | most | [few,most] | most | |
| all | most | all | most | |
| most | most | all | most | [half,most] |
| most | all | [few,all] | most | |
| half | [few,all] | all | all | [half,half] |
| all | all | [few,all] | half | |
| most | [few,all] | [few,most] | all | [most,all] |
| most | [few,all] | all | all | [most,most] |
| all | all | [few,all] | most | |
| all | [few,all] | [few,all] | all | [all ,all] |

Table 1: Compacted table of the quantified syllogism for the 5 element partition ('half' means 'about half')

The case P(B|A) = 'all' and P(B|C) = 'all' represents a typical case of statistical inference, when, knowing the probability P(C|B), and considering some individual in class B, one tries to say something about its probability of being a C. Namely B represents a population, C a subclass of this population for which the proportion or probability P(C|B) is known. For instance B represents the inhabitants of some city and C the proportion of individuals in that population that are older than 60. Now take an individual $x_0$ in B. There are several ways of considering $x_0$ according to its peculiarities. Let A be the maximal subset of B containing individuals "just like $x_0$". Note that A can range from $\{x_0\}$ (if $x_0$ is so particular as nobody is like him) to B itself (if $x_0$ is viewed as having nothing special). The problem is then : knowing P(C|B) what is the probability that $x_0$ belongs to C ? This problem can be solved by computing P(C|A) where A is a maximal subclass of B, of which $x_0$ is a typical element.

This problem corresponds to all rows of Table 1 where P(B|A) = 1 and P(B|C) = 1. It can be checked that P(C|A) can be much more imprecise than P(C|B), since it can be [none, all] (i.e."unknown") in several cases.

This phenomenon can be precisely studied in an analytical way, letting P(C|B) = α, and P(A|B) = t. Parameter t can be called a typicality index of set A with respect to B. It expresses the probability that selecting at random an individual in B, it lies in A, i.e. it is "like $x_0$". The commonsense saying that statistics should be cautiously used when making decisions about individual situations can be given a precise form thanks to the quantified syllogism. When P(A|B) = t, P(B|A) = 1, P(C|B) = α, P(B|C) = 1, we get the following results on P(C|A) :

$$P_*(C|A) = \max\left(0, 1 - \frac{1-\alpha}{t}\right) ; P^*(C|A) = \min\left(1, \frac{\alpha}{t}\right)$$

The only case when P(C|A) can only be equal to P(C|B) is when t = 1, i.e. when the reference class of $x_0$ is B itself. Let us consider the situation where P(C|B) > 1/2. If the degree of typicality t ≤ P(C|B) = α then the probability P(C|A) is no longer upper bounded, but can be lower than P(C|B) as well. When the typicality t is low enough, that is t ≤ min(P(C|B), 1 – P(C|B)) nothing can be inferred on P(C|A). It corresponds to the case when A and C could be disjoint subsets of B. This phenomenon explains the presence of rows of Table 1 where despite the high values of some of the probabilities the results of the chaining is very imprecise.

## 5   ROBUSTNESS ANALYSIS

Table 1 is obtained for a specific value of the threshold α between "few" and "half". A legitimate question is whether such results are still valid for other values of the threshold. Let us start with qualitative tables for product and quotient, with "few" = (0,α], "most" = [1 – α, 1), "half" = [α, 1 – α]. The product table is defined as None · Q = None, All · Q = Q, and

| · | few | half | most |
|---|---|---|---|
| few | few | few | few |
| half | few | ? | [few, half] |
| most | few | [few, half] | ? |

The question mark '?' indicates some ambiguity due to the choice of the value of α. Namely half * half = [α², (1 – α)²] ⊆ (0, α] only if (1 – α)² ≤ α, which requires α ≥ d, where $d = \frac{3 - \sqrt{5}}{2} \approx 0.382$. In that case half * half = few and most * most = [(1 – α)², 1) ⊄ [α, 1) when α > d, so that most * most = [few, most]. The latter equality does not sound natural. On the contrary if α < d, then half * half = [few, half] ; most * most = [half, most]. From a commonsense point of view, it is not very unnatural to require that "few" may mean a proportion less than .3 or so. Again "half" is here short for "neither few nor most but in-between". Hence it is clear that the product of qualitative probabilities is almost independent of the choice of the threshold α in (0, 1/2). It fits the intuition and is completely threshold-independent for "α" small enough. The same problem can be solved for the (bounded) quotient, and it leads to the following almost-robust table

| / | none | few | half | most | all |
|---|---|---|---|---|---|
| none | [none, all] | none | none | none | none |
| few | all | [none, all] | [few, all] | [few, half] | few |
| half | all | all | [half, all] | [half, all] | half |
| most | all | all | all | most | most |
| all | all | all | all | all | all |

The terms half/half and few/most are given for $\alpha < d$. Only these terms change if $\alpha$ is larger. Note that the subdiagonal part of the table has been truncated to 1.

In order to study the robustness of the quantified syllogism table, several runs of the program that generate this table have been done, with $\alpha$ varying between 0.25 and .38. Only a few lines of the qualitative table change (nine over $625 = 5^4$ distinct 4-tuples of quantifiers for $.025 \leq \alpha \leq .35$). In order to get a better insight, it is interesting to consider a significant subpart of the table, where quantifiers are either "few" or "most", i.e. when P(A|B), P(B|A), P(C|B), P(B|C) are close to 0 or close to 1. In order to let the parameter $\alpha$ appear we shall use the following notation

$$P(A|B) \; V_0 (\alpha) \text{ which means } P(A|B) \leq \alpha$$

$$P(A|B) \; V_1 (\alpha) \text{ which means } P(A|B) \geq \alpha.$$

Then by applying the optimal bounds on P(C|A) as described in Section 3 on the $16 = 4^2$ 4-tuples of extreme quantifiers, potential instability of the results was obtained for the 6 following cases only :

| P(B|A) | P(A|B) | P(B|C) | P(C|B) | P(C|A) |
|---|---|---|---|---|
| $V_0(\alpha)$ | $V_1(1-\alpha)$ | $V_1(1-\alpha)$ | $V_0(\alpha)$ | $V_0(\alpha^2/(1-\alpha)^2)$ |
| $V_0(\alpha)$ | $V_1(1-\alpha)$ | $V_1(1-\alpha)$ | $V_1(1-\alpha)$ | $V_0((\alpha^2/(1-\alpha)^2)+\alpha)$ |
| $V_1(1-\alpha)$ | $V_1(1-\alpha)$ | $V_0(\alpha)$ | $V_0(\alpha)$ | $V_0(2\alpha)$ |
| $V_1(1-\alpha)$ | $V_1(1-\alpha)$ | $V_0(\alpha)$ | $V_1(1-\alpha)$ | $V_1(1-2\alpha)$ |
| $V_1(1-\alpha)$ | $V_1(1-\alpha)$ | $V_1(1-\alpha)$ | $V_0(\alpha)$ | $V_0(\alpha/((1-\alpha)^2+\alpha^2))$ |
| $V_1(1-\alpha)$ | $V_1(1-\alpha)$ | $V_1(1-\alpha)$ | $V_1(1-\alpha)$ | $V_1(1-2\alpha)$ |

Table 2

It is easy to verify that for $\alpha \leq 1/3$

$$\frac{\alpha^2}{(1-\alpha)^2} \leq \alpha \qquad \alpha + \frac{\alpha^2}{(1-\alpha)^2} \leq 1 - \alpha$$

$$2\alpha \leq 1 - \alpha \qquad \alpha \leq 1 - 2\alpha$$

$$\frac{\alpha}{(1-\alpha)^2 + \alpha^2} \leq 1 - \alpha \; \text{(since} \; \frac{\alpha}{(1-\alpha)^2+\alpha^2} \leq 2\alpha).$$

These inequalities guarantee that whatever the value of $\alpha \leq 1/3$, the value of P(C|A), as shown in Table 2 remains within a given range (e.g. $(0, \alpha]$, $(0, 1-\alpha]$, $[\alpha, 1))$ corresponding to a symbolic label, even if there is a degradation of the result which is less specific than $V_0(\alpha)$ or $V_1(1-\alpha)$ (except in the first line of Table 2). Hence we get the following robust computation table for the quantified syllogism (we only give here the 4-tuples that lead to an informative output) :

| P(B|A) | P(A|B) | P(B|C) | P(C|B) | P(C|A) |
|---|---|---|---|---|
| few | most | few | few | [none, most] |
| few | most | most | few | few |
| few | most | few | most | [few, all] |
| few | most | most | most | [few, half] |
| most | most | few | few | [none, half] |
| most | most | few | most | [half, all] |
| most | most | most | few | [none, half] |
| most | most | most | most | [half, all] |

When the quantifier "(about) half" is involved in the inference pattern, the resulting quantifiers P(C|A) may get more precise (e.g. [few, all] becomes [half, all] when $\alpha$ becomes smaller). But the table obtained for $\alpha = 0.3$ remains correct but not optimally precise.

## 6  A QUALITATIVE ANALYSIS OF ADAMS' INFERENCE RULES

Adams (1975) has proposed a probabilistic inference system based on the three inference rules :

triangularity :   $A \to B, A \to C \Rightarrow (A \land B) \to C$
Bayes rule :       $A \to B, (A \land B) \to C \Rightarrow A \to C$
disjunction :      $A \to C, B \to C \Rightarrow (A \lor B) \to C$

which are sound when $A \to B$ is understood as the probability $P(B|A) \geq 1 - \varepsilon$ where $\varepsilon$ is arbitrarily small. These rules are used in Pearl (1988) to build a probabilistic inference-like default logic. It is interesting to consider finistic semantics for these rules in relationship with the linguistic probability scale. In this respect $A \to B$ will be interpreted as "most A's are B's". First it is easy to verify that triangularity and Bayes rule axioms can be expressed in terms of the quantified syllogism, of which they are special cases, noticing that $P(B|A) = P(A \cap B|A)$.

Triangularity : $P(A \cap B|A) =$ most ; $P(A|A \cap B) = 1$ ; $P(C|A) =$ most; compute $P(C|A \cap B)$;

Bayes rule : $P(A|A \cap B) = 1$ ; $P(A \cap B|A) =$ most ; $P(C|A \cap B) =$ most ; compute $P(C|A)$.

Taking 'most' $= [1 - \alpha, 1)$, we easily get the lower bound on $P(C|A \cap B)$ and $P(C|A)$ in each case by using the quantified syllogism

$$P(C|A \cap B) \geq \max\left(0, 1 - \frac{1 - P_*(C|A)}{P_*(A \cap B|A)}\right) = \frac{1 - 2\alpha}{1 - \alpha}$$

$$P(C|A) \geq P_*(B|A) \cdot P_*(C|A \cap B) = (1-\alpha)^2.$$

There is again a degradation of the lower bounds. However these lower bounds are again greater than $\alpha$ when $\alpha \leq d$.

The third axiom pertains to another kind of inference that does not directly relate to the quantified syllogism. In Amarger et al. (1991a) the following identity was obtained :

$$P(C|A \cup B) = \frac{\dfrac{P(C|A)}{P(B|A)} + \dfrac{P(C|B)}{P(A|B)} - P(C|A \cap B)}{\dfrac{1}{P(B|A)} + \dfrac{1}{P(A|B)} - 1}$$

Hence a lower bound to $P(C|A \cup B)$ is obtained when $P(C|A \cap B) = 1$. When $P(C|A) \geq 1 - \alpha$, $P(C|B) \geq 1 - \alpha$ (both express "most"), we get

$$P(C|A \cup B) \geq \frac{K(1-\alpha) - 1}{K - 1}$$

where $K = \frac{1}{P(B|A)} + \frac{1}{P(A|B)} \geq 2$. The right-hand term of the inequality is increasing with K. Hence the lower bound for $P(C|A \cup B) \geq 2(1-\alpha) - 1 = 1 - 2\alpha$. More generally $P(C|A \cup B) \geq 1 - \alpha - \alpha'$ when $P(C|A) \geq 1 - \alpha$, $P(C|B) \geq 1 - \alpha'$. On the whole, we have found finistic counterparts of Adams' axioms that enable to quantify how inaccurate we are when we apply these axioms for commonsense reasoning with high probabilities.

The three axioms can be summarized as

$$A \xrightarrow[\alpha]{} B, A \xrightarrow[\alpha]{} C \Rightarrow A \cap B \xrightarrow[1-\alpha]{\alpha} C$$

$$A \xrightarrow[\alpha]{} B, A \cap B \xrightarrow[\alpha]{} C \Rightarrow A \xrightarrow[2\alpha - \alpha^2]{} C$$

$$A \xrightarrow[\alpha]{} C, B \xrightarrow[\alpha]{} C \Rightarrow A \cup B \xrightarrow[2\alpha]{} C$$

where $A \xrightarrow[\alpha]{} B$ reads $P(B|A) \geq 1 - \alpha$. In terms of linguistic proportions, those rules can be written changing $\alpha$ into "most" and interpreting the resulting conditional probabilities as "more than few" in the three cases, provided that $\alpha < 1/3$. These rules enable probabilistic reasoning to be performed as a qualitative non-monotonic logic, but where the validity of conclusions can be numerically assessed.

## 7 THE GENERALIZED BAYES THEOREM

In the case of the generalized Bayes theorem (GBT), described in Section 3, we cannot use the same method as we did in Section 4 with the quantified syllogism rule because here the number of arguments, i.e. the length of the involved cycle, is variable. This prevents us from having the qualitative inference pattern defined by a table. Then the only possibility left is to replace in the GBT expression the product and quotient operations by qualitative ones defined on the universe of description U. These more basic qualitative operations can be stored in tables.

Given a cyle $(A_1, ..., A_k, A_1)$ with $A1 = A$, $A_k = B$ the qualitative probability QP(A|B), known to lie in the interval $[QP_*(A|B)_{old}, QP^*(A|B)_{old}]$ should be improved by letting

$$QP_*(A|B)_{new} =$$

$$QP_*(A|B)_{old} \vee \left[ QP_*(B|A) \cdot \frac{\prod_{i=1}^{k-1} QP_*(A_i|A_{i+1})}{\prod_{i=1}^{k-1} QP^*(A_{i+1}|A_i)} \right]$$

$$QP^*(A|B)_{new} =$$

$$QP^*(A|B)_{old} \wedge \left[ QP^*(B|A) \cdot \frac{\prod_{i=1}^{k-1} QP^*(A_i|A_{i+1})}{\prod_{i=1}^{k-1} QP_*(A_{i+1}|A_i)} \right]$$

$\wedge$ and $\vee$ denote the min and max operations in the sense of the certainty ordering. But the computation of these quantities raises several problems

i) for a given cycle, find a proper ordering for the computation. Especially, it is not obvious that $(X1 \cdot X2) / (X3 \cdot X4)$ (computing products first) is equal to $(X1/X2) \cdot (X3/X4)$ (computing quotients first). Because of the truncation effect of the quotient table, it seems better to compute products before quotients.

ii) since this operation must be done for all cycles one might look for the counterpart of a longest path algorithm, here with qualitative values. But this is tricky if we want to compute quotients only at the end of the shortest path procedure, and keep separate the products of terms along cycles. The maximum operation $(Q_1/Q_2) \vee (S_1/S_2)$ should be directly expressed as an operation $\vee'$ between pairs $(Q_1,Q_2)$ and $(S_1,S_2)$ that furnishes a new pair of qualitative values. Moreover, longest path algorithms make an extensive use of the distributivity of the addition over the maximum. Here we would require a property such as

$$\frac{X_1}{X_2} \vee \left( \frac{X_3 \cdot X_4}{X_5 \cdot X_6} \right) = \frac{X_1 \cdot X_3}{X_2 \cdot X_5} \vee \frac{X_1 \cdot X_4}{X_2 \cdot X_6}$$

It is not clear that this property holds in the qualitative algebra.

But the basic question is whether this constraint propagation rule, which proved useful in the quantitative case leads to really improve qualitative probability bounds. This can be precisely studied on the 5 quantifier case of Section 4.2.

The smallest expression to be computed with non-extreme probabilities is of the form $(Q_1 \cdot Q_2 \cdot Q_3)/(Q_4 \cdot Q_5)$ with $Q_i \in \{few, half, most\}$. It is easy to check from the product and quotient tables that

i) $Q_1 \cdot Q_2 \in \{few, [few, most], [few, half], [half, most]\}$

ii) $Q_1 \cdot Q_2 \cdot Q_3$ can only belong to the same set as above

iii) the only case where a quotient can be significantly informative is when the operands are [few, most], [half, most] and [half, half] since few/most = [few, half] and half/half = [half, all] = half/most.

As a consequence $(Q_1 \cdot Q_2 \cdot Q_3)/(Q_4 \cdot Q_5)$ can give [few, all] at the very best. This is when $Q_1 \cdot Q_2 \cdot Q_3 =$ [few, half] and $Q_4 \cdot Q_5 =$ [half, most]. This is not likely to be very useful for improving probability bounds. In the 7-quantifier case, the best informative result can be shown to be [half, all] corresponding to when

$$\frac{Q_1 \cdot Q_2 \cdot Q_3}{Q_4 \cdot Q_5} = \frac{[half, most]}{[half, most]}, \text{ and } \frac{[most, al-all]}{[most, al-all]}.$$

## 8  SYMBOLIC CONSTRAINT PROPAGATION

The quantified syllogism rule, as precomputed for a given linguistic partition of the unit interval, can be recursively applied to any set of linguistic quantified statements of the form $Q$ $A_i$'s are $B_j$'s which form a probabilistic network. It is then possible to generate new statements of that kind, and to improve precision for the ones that were originally stated. Let us consider the following qualitative counterpart of a 5-predicate example of Amarger et al. (1991a, b) :

- Most to almost all students are sportsmen (Q = [most, al-all])
- Almost all students are young (Q = al-all)
- Half of the sportsmen are students (Q = half)
- Almost all sportsmen are single (Q = al-all)
- At least almost all sportsmen are young (Q = [al-all, all])
- At least most singles are sportsmen (Q = [most, all])
- Most singles are young (Q = most)
- Almost no singles have children (Q = al-none)
- Few young people are students (Q = few)
- Almost all young people are sportsmen (Q = al-all)
- At most almost no young people have children (Q = [none, al-none])
- At most almost no people who have children are single (Q = [none, al-none])
- At most almost no people who have children are young (Q = [none, al-none])

These statements are but examples and must not be examined as to their actual truthworthiness.

Let us consider a 7-element partition as follows

partition : (0, 0.2, 0.4, 0.6, 0.8, 1)
$\mathcal{P}$ = {none, al-none, few, half, most, al-all, all}.

The quantified syllogism rule is run until no improvement of the quantifiers, nor new statements can be generated. The following results were obtained :

- At least few students are single (Q = [few, all])
- Not more than few sportsmen have children (Q = [none, few])
- From almost-none to half singles are students (Q = [al-none, half]).

Let us consider now a 9-element partition as follows

partition : (0, 0.1, 0.2, 0.4, 0.6, 0.8, 0.9, 1)
$\mathcal{P}$ = {none, al-none, v-few, few, half, most, v-many, al-all, all}

where v = few stands for very-few ([0,1, 0.2]) and v-many stand for very many ([0.8, 0.9]). Using the same input data, more results are obtained :

- At least half of the students are single (Q = [half, all])
- Not more than half of the students have children (Q = [none, half])
- Not more than very few sportsmen have children (Q = [none, v-few])
- Most to very many singles are sportsmen (Q = [most, v-many]).

These results are consistent but stronger than those obtained with the 7-element partition. It is interesting to compare them with the results of the numerical procedure that directly handles interval probabilities given below under the form of incidence matrices giving P(<column>|<row>) :

| input data | student | sport | single |
|---|---|---|---|
| student | [1.000,1.000] | [0.700,0.900] | [0.000,1.000] |
| sport | [0.400,0.600] | [1.000,1.000] | [0.800,0.850] |
| single | [0.000,1.000] | [0.700,0.900] | [1.000,1.000] |
| young | [0.250,0.350] | [0.800,0.900] | [0.900,1.000] |
| children | [0.000,1.000] | [0.000,1.000] | [0.000,0.050] |

| input data | young | children |
|---|---|---|
| student | [0.850,0.950] | [0.000,1.000] |
| sport | [0.900,1.000] | [0.000,1.000] |
| single | [0.600,0.800] | [0.050,0.800] |
| young | [1.000,1.000] | [0.000,0.050] |
| children | [0.000,0.050] | [1.000,1.000] |

Input data

| saturated network | student | sport | single |
|---|---|---|---|
| student | [1.000,1.000] | [0.900,0.900] | [0.607,1.000] |
| sport | [0.400,0.400] | [1.000,1.000] | [0.850,0.850] |
| single | [0.222,0.366] | [0.700,0.700] | [1.000,1.000] |
| young | [0.350,0.350] | [0.834,0.888] | [0.900,0.900] |
| children | [0.000,0.099] | [0.000,0.127] | [0.000,0.050] |

| saturated network | young | children |
|---|---|---|
| student | [0.850,0.850] | [0.000,0.271] |
| sport | [0.900,0.958] | [0.000,0.154] |
| single | [0.800,0.800] | [0.050,0.100] |
| young | [1.000,1.000] | [0.000,0.050] |
| children | [0.000,0.044] | [1.000,1.000] |

Probabilities after constraint propagation

The main difference between the numerical and the symbolic results appears on the last row. The symbolic inference approach was not able to deduce that almost nobody having children is a student, and very few are sportsmen. Note that we have tried to develop a qualitative version of the generalized Bayes rule using longest path algorithms and the product and quotient tables of computation. However no improvement of the results has been observed. More work is to be done along that line.

## 9 CONCLUDING REMARKS

We have shown in this paper that a qualitative calculus for the probabilistic scale 'none', 'few', 'from few to most', 'most', 'all' can be developed in agreement with a numerical interpretation of probabilities, provided that the intended numerical meaning of 'few' is less than 33 % in any case and the one of 'most' is more than 66 %. These thresholds are quite in agreement with commonsense which seems to disagree that "most A's are B's" if less than 70 % of A's are B's, or that "few A's and B's" when there are more than 30 % of A's which are B's. However it does not mean that humans are currently able to provide the correct (in the sense of probability calculus) qualitative values given by the rules derived in this paper. It is well known (e.g. Kahneman, Slovic and Tversky, 1980) that humans are often in trouble not only for correctly assessing probabilities, but also to make accurrate inference from them.

One might wonder whether fuzzy intervals are useful or not in the modeling of linguistic quantifiers. Clearly the use of precise thresholds to delimit the extensions of "few", "half", "most" has something arbitrary. However since the linguistic computation tables obtained here are partially independent of the choice of the threshold, it turns out that using fuzzy partitions instead of non-fuzzy ones would not make much difference here, especially if a fuzzy partition is viewed as an imprecise specification of the thresholds between the meanings of the basic terms. Nevertheless fuzzy intervals remain useful in the scope of feeding numbers in probabilistic networks, from the knowledge of linguistic values, rather than reasoning *with* linguistic values. Indeed, when looking for the numerical interpretation of linguistic quantifiers, fuzzy intervals look like a more faithful model than crisp ones. But then the constraint propagation algorithms must be adapted to handle fuzzy upper and lower probabilities in the numerical setting. Applying fuzzy arithmetic to the quantified syllogism rule (as done by Dubois and Prade (1988)) appears to be in total contrast with defining linguistic counterparts of numerical constraint propagation rules, as done here.

**Acknowledgments** : This work is partially supported by the ESPRIT-Basic Research project n°3085, DRUMS.